\documentclass[11pt]{article}
\usepackage{acl2016}
\usepackage{times}
\usepackage{url}
\usepackage{latexsym}
\usepackage{amsmath}

\usepackage{graphicx}
\usepackage{todonotes}
\usepackage{amssymb}

\usepackage{pifont}
\newcommand{\cmark}{\ding{51}}%
\newcommand{\xmark}{\ding{55}}%

\usepackage{color}

\aclfinalcopy 

\newcommand{\Note}[2]{} 
\newcommand{\SideNote}[2]{} 
\renewcommand{\Note}[2]{\todo[color=#1,size=\small, inline=true]{#2}} 
\renewcommand{\SideNote}[2]{\todo[color=#1,size=\small]{#2}} 

\title{Feature Generation for Robust Semantic Role Labeling}

\author{Travis Wolfe\qquad Mark Dredze\qquad Benjamin {Van Durme} \\
  Human Language Technology Center of Excellence \\
  Johns Hopkins University}

\date{}

\begin{document}
\maketitle
\begin{abstract}
	Hand-engineered feature sets are a well understood method for
	creating robust NLP models, but they require a lot of expertise
	and effort to create.
	In this work we describe how to automatically generate rich feature sets from
	simple units called featlets, requiring less engineering.
	Using information gain to guide the generation process,
	we train models which rival the state of the art on two
	standard Semantic Role Labeling datasets with
	almost no task or linguistic insight.
\end{abstract}



\section{Introduction}
Feature engineering is widely recognized as an important component of
robust NLP systems,
with much of this engineering done by hand.
Articles describing improvements in task performance over prior work tend to be
methodologically driven
(for example low-regret online learning algorithms and structured regularizers),
with improvements in feature design
often described just briefly, and as a matter of secondary importance.  While
the distinctions between methods of inference are formalized in the language of
mathematics, most expositions of feature design employ terse, natural language
descriptions, often not sufficient for reliable reproduction of the underlying
factors being extracted.  This has led to stagnation in feature design, and in
general an attitude in some circles that features themselves are not worth
exploring; i.e., we should abandon explicit, interpretable features for neural
techniques
which create their own representations which may not align with our own.


Features sets are constructed by
authors using heuristics which are often not tested. For example it is common
to coarsen a feature before using it in a product because the fine grained product
would produce ``too many" features.
The author may have been correct (they ran the experiment and verified that performance
went down) or not, but the reader often doesn't know which is the case,
and are left with the same problem of whether to run that experiment or not.
Due to the cost of running experiments, practitioners are biased towards copying
the feature set verbatim.


This work is about removing the human from the loop of feature selection,
focussing on Semantic Role Labeling (SRL) \cite{gildea2002automatic}.
The key challenge that we address is feature generation.
Previous work has generated features by taking the Cartesian product of
templates, but this is not rich enough to capture many widely used
manually created features.
We show that by decomposing the template even further, into atoms called
featlets, we can automatically compose templates with rich,
ad-hoc combinators.
This process can generate many features which
an expert might not consider.

Once we have tackled the feature generation problem,
we show that we can automatically derive feature sets
which match the performance of state-of-the-art feature sets created by experts.
Our method uses basic statistics and requires no human expertise.
We believe that models specified using featlets are easier to reproduce
and offer the potential for performing feature selection with
machine learning rather than domain expert knowledge,
potentially at lower-cost and super-human performance.

\paragraph{Feature Descriptions in the Literature}
For a case study
on feature descriptions,
consider the ``voice feature" for SRL.
It was first motivated and described in \newcite{gildea2002automatic}.
They said that they defined 10 rules for when a verb had either
active or passive voice, but never said what they were.
Since then, almost every prominent paper on SRL has listed voice as
a feature or template that they use, but none of the following
defined their rules for the voice feature.\footnote{
\newcite{gildea2002automatic}
\newcite{xue2004calibrating}
\newcite{pradhan2005support}
\newcite{Toutanova:2005}
\newcite{Johansson:2008}
\newcite{Marquez:2008}
\newcite{punyakanok2008importance}
\newcite{DasFramesCL:2014}}
Further, discrepancies between authors is not unheard
of: \newcite{gildea2002automatic} report 5\% of verbs
were passive in the PTB, while \newcite{pradhan2005support} report 11\%.
Some of these papers go into great detail about other aspects
like ILP constraints and
and regularization constants,
but this same clarity doesn't always extend to features.

In methods papers, math is used as a bridge
between natural and programming languages.
There is no equivalent for describing features,
so this type of omission is understandable
given space constraints and the clumsiness of natural language.
However, given the importance of the underlying factors in a model, the lack of clarity
diminishes the value of the work to other practitioners, especially among
those less linguistically-inclined.

\section{Featlets}
Our approach begins with the notion that features can be decomposed into smaller
units called featlets.
These units can be composed together to make a wide variety of features.
We distinguish featlets from {\em feature templates}, or just {\em templates},
which are effectively sets of features. Featlets are not necessarily features,
but are composed to produce features or feature templates.


To start with an example, the featlet \textsc{Word}: given the index
of a token, it returns the word at that position. A feature would not assume
a token index is given, only that $y$ and $x$ are given, so \textsc{Word} is not a feature.
Featlets are also used to provide information to other featlets.
For example, the featlet \textsc{ArgHead} takes the head token index of an argument span
and passes it to \textsc{Word}.
The combination of the two, {\small [\textsc{ArgHead}, \textsc{Word}]}, is a template.
Importantly featlets are interchangeable: the template {\small [\textsc{ArgHead}, \textsc{Word}]}
is related to {\small [\textsc{ArgHead}, \textsc{Pos}]} and
{\small [\textsc{ArgHeadParent}, \textsc{Word}]}.
Featlets are minimal to ensure that the trial and error
of feature engineering falls to the machine rather than the expert.

\paragraph{Definition}
\label{sec:featlets-formal}
Featlets are operations performed on
a {\em context},
which is a data structure with:
\begin{enumerate}
\itemsep0em
\item Named fields which have types 
\item A list of featlets which have been applied 
\item An output buffer of features 
\end{enumerate}

In our implementation the data fields are:
\begin{itemize}
\itemsep0em
\item \texttt{token1} and \texttt{token2}: are integers
\item \texttt{span1} and \texttt{span2}: are pairs of integers (start, end)
\item \texttt{value}: an untyped object 
\item \texttt{sequence}: is an untyped list
\end{itemize}

Each of the fields in a context start out as a special value \textsc{Nil}.
Once they are set, other featlets can read from these fields
and put a feature into the output buffer.
If a \textsc{Nil} field is read, then the featlet fails and no features
are output.


\paragraph{Label Extractors}
\label{sec:label-extractors}
This group of featlets are responsible for
reading an aspect of the label $y$ and putting it into the context.
These are the only task-specific featlets which the inference algorithm
has to be aware of.
\begin{itemize}
\itemsep0em
\item \textsc{TargetSpan}: sets \texttt{span2} to a target
\item \textsc{TargetHead}: sets \texttt{token2} to the head of the target span
\item \textsc{ArgSpan}: sets \texttt{span1} to an argument span
\item \textsc{ArgHead}: sets \texttt{token1} to the head of an argument span
\item \textsc{Role}: sets \texttt{value} to a role
\item \textsc{FrameRole}: sets \texttt{value} to the concatenation of a frame and a role
\item \textsc{Frame}: sets \texttt{value} to a frame
\end{itemize}

\paragraph{Token Extractors}
These read from \texttt{token1} and output a feature.
\begin{itemize}
\itemsep0em
\item \textsc{Word}, \textsc{Pos}, \textsc{Lemma}
\item \textsc{WnSynset}: reads the lemma and POS tag at \texttt{token1}, looks up the
	first WordNet sense, and puts its synset id onto \texttt{value}.
\item \textsc{BrownClust}: looks up a un-supervised hierarchical word cluster id for \texttt{token1}\footnote{One featlet for
	a 256 and a 1000 cluster output of \newcite{liang2005semi}.}
\item \textsc{DepRel}: output the syntactic dependency edge label connecting \texttt{token1}
  to its parent.
\item \textsc{DepthD}: compute the depth of \texttt{token1} in a dependency tree
\end{itemize}

Before moving on, it will be helpful to slightly redefine the behavior of
token extractors: instead of immediately outputting a string, instead they
will store that string in the \texttt{value} field and leave it to the
\textsc{Output} featlet to finish the job of outputting a feature.
By convention, we will assume that every string of featlets ends in
a (possibly implicit) \textsc{Output}, so the old meaning of ``token extractors
output a feature" is true as long as the token extractor featlet is
not followed by anything.

\paragraph{Value Mutators}
In many cases though, we will want normalized or simplified versions of
other features.
For example we could
want to find the shape of a word, ``McDonalds" $\rightarrow$ ``CcCcccccc", or perhaps
just take the first few characters, ``NNP" $\rightarrow$ ``N".
Value mutators read a string from \texttt{value}, compute a new string, and store it back to \texttt{value}.
This enables features like {\small [\textsc{ArgHead}, \textsc{Word}, \textsc{Shape}]}
or {\small [\textsc{TargetHead}, \textsc{Pos}, \textsc{Prefix1}]}.
\begin{itemize}
\itemsep0em
\item \textsc{LC}: if value is a string, output its lowercase
\item \textsc{Shape}: if value is a string, output its shape
\item \textsc{PrefixN}: sets \texttt{value} to a prefix of length 
\end{itemize}

An interesting special case of value mutators are ones which filter.
A featlet like \textsc{ClosedClass} can be applied after \textsc{Word}
but before \textsc{Output} in order to have a feature only fire for closed
class words. This is achieved by \textsc{ClosedClass} writing \textsc{Nil}
to \texttt{value} so that \textsc{Output} fails and no features are output.
This selective firing is valuable because it can lead to
expressive feature semantics (e.g. ``only output the first word in a span
if is in a closed class"). 

\paragraph{Dependency Walkers}
Syntax lets us jump around in a sentence where structural proximity is often a more
informative measure of relevance than linear proximity.
Dependency walkers\footnote{Every time we list a \textsc{Left} featlet, we have omitted
its \textsc{Right} equivalent for space.} provide one way of jumping around by reading and writing
\texttt{token1}.
These can be composed as well to form walks,
e.g. ``grandparent" = {\small [\textsc{ParentD}, \textsc{ParentD}]}.
\begin{itemize}
\itemsep0em
\item \textsc{ParentD}: set \texttt{token1} to its parent in the dependency tree.
\item \textsc{LeftChildD}: sets \texttt{token1} to its left-most child in the dependency tree
\item \textsc{LeftSibD}: sets \texttt{token1} to its next-to-the-left sibling in the dependency tree
\item \textsc{LeftMostSibD}: sets \texttt{token1} to its left-most sibling in the dependency tree
\end{itemize}

Some information is contained in the name of a dependency walker (e.g. \textsc{ParentD}),
other information is contained in the edge crossed (e.g. whether the parent is
\texttt{nsubj} or \texttt{dobj}).
To capture this information, dependency walkers also append the edge that they crossed into the
\texttt{sequence} field.
This side information can be read out later by other featlets.

\paragraph{Sequence Reducers}
Values appended to \texttt{sequence} are converted into features
with sequence reducers.
\begin{itemize}
\itemsep0em
\item \textsc{Ngrams}: reads n-grams from sequence, outputs each.
  If \texttt{value} is set, prefixes every n-gram with \texttt{value}.
  Clears \texttt{sequence} when done.
\item \textsc{Bag}: special case of n-grams when n=1
\item \textsc{SeqN}: if sequence is no longer than N, output items concatenated (preserves order).
  Also will prepend \texttt{value} if set.
  Clears \texttt{sequence} regardless of length.
\item \textsc{CompressRuns}: Collapses X Y Y Z Z Z to X Y+ Z+, no output, doesn't clear \texttt{sequence}
\end{itemize}

\paragraph{Dependency Representers}
Going back to dependency walkers for a moment,
the edge that they append to \texttt{sequence}
need not be a string like \texttt{nsubj}
which sequence reducers can operate on.
Edges are represented as tuples of (parent, child, deprel)\footnote{Sequence
reducers fail when attempting to operate non-string values in \texttt{sequence},
so a representer must be called first.}, and
we add featlets which choose a string to represent each edge,
called \textbf{dependency representers}.
We construct an edge to string function by taking the Cartesian
product of the token extractors to represent the parent and
the set of functions
$\{$
\textsc{EdgeDirectionD}\footnote{Left or right},
\textsc{EdgeLabelD}\footnote{Taken from dependency parse, e.g. \texttt{nsubj}},
\textsc{NoEdgeD}\footnote{A constant string: ``*"}
$\}$
and make a featlet to map each of these functions over \texttt{sequence}.

\paragraph{Constituent Walkers}
There is an equivalent class of featlets to the dependency walkers which
instead operate on \texttt{span1}, \texttt{span2}, and a constituency tree,
called constituent walker.
Each of these operations fail if the span they read is not a constituent.
\begin{itemize}
\itemsep0em
\item \textsc{ParentC}: sets \texttt{span1} to its parent
\item \textsc{LeftChildC}: sets \texttt{span1} to the left-most child node
\item \textsc{LeftSibC}: sets \texttt{span1} to the next-to-the-left sibling node
\item \textsc{LeftMostSibC}: sets \texttt{span1} to the left-most sibling node
\end{itemize}

\paragraph{Constituent Representers}
Constituent walkers differ from dependency walkers in the values
that they append to \texttt{sequence},
they store grammar rules like \mbox{S $\rightarrow$ NP VP}.
Equivalently to dependency representers, the constituency representers are
\mbox{$\{$ \textsc{CategoryC}\footnote{The left hand side of a rule},
\textsc{SubCategoryC}\footnote{The left and right hand side of a rule} $\}$}.

\paragraph{Tree Walkers}
Operations longer than one step typically require that a start and endpoint
are known to avoid meaningless walks.
Tree walkers use both dependency
and constituency parses,
but take shortest path walks between two endpoints,
adding edges or rules to \texttt{sequence}.
\begin{itemize}
\itemsep0em
\item \textsc{ToRootD}: walks from \texttt{token1} to root
\item \textsc{CommonParentD}: walks from \texttt{token1} to a common parent and
then \texttt{token2}
\item \textsc{ToRootC}: walks from \texttt{span1} to root
\item \textsc{CommonParentC}: like \textsc{CommonParentD} for constituency trees
\item \textsc{ChildrenD}: walks the children of \texttt{token1}, left to right
\item \textsc{ChildrenC}: walks the children of \texttt{span1}, left to right
\end{itemize}

\paragraph{Linear Walkers}
If syntactic trees are not available or accurate,
linear walkers can provide another source of relevant information.
These featlets append token indices to \texttt{sequence}
for multi-step walks, but behave
like dependency walkers otherwise, mutating a field such as \texttt{token1}.
\begin{itemize}
\itemsep0em
\item \textsc{LeftL}: moves \texttt{token1} position if possible
\item \textsc{Span1StartToEndL}: walks tokens in \texttt{span1}
\item \textsc{Span1LeftToRightL}: like \textsc{Span1StartToEndL} but expands two tokens in either direction
\item \textsc{Head1ToSpan1StartL}
\item \textsc{Head1ToSpan1EndL}
\item \textsc{Span1ToSpan2L}: adds any tokens between the two spans to \texttt{sequence}
\end{itemize}

\paragraph{Distance Functions}
Distance can be informative, but usually not clear how to represent its scale.
Featlets let us address which distances to measure separately
from the units used to measure them.
Distance functions put a number into the \texttt{value} field:
\begin{itemize}
\itemsep0em
\item \textsc{SeqLength}: the number of elements in \texttt{sequence}\footnote{This
	can be use to measure a variety of distances using linear walkers, like \texttt{ArgSpan} width
	or the distance between \texttt{ArgHead} and \texttt{TargetHead}.}
\item \textsc{DeltaDepthD}: if values in \texttt{sequence} are dependency nodes or token indices,
	put the	depth of the first minus the second into \texttt{value}
\item \textsc{DeltaDepthC}: like \textsc{DeltaDepthD} for constituency nodes.
\end{itemize}

\paragraph{Distance Representers}
Once a number has been put into \texttt{value},
distance representers write a string representation
suitable for a feature back to \texttt{value}.
\begin{itemize}
\itemsep0em
\item \textsc{DasBuckets}: encodes the bucket widths defined in \newcite{DasFramesCL:2014}
\item \textsc{Direction}: writes \texttt{+1} or \texttt{-1} based on the sign
	of the number given.
\end{itemize}

\subsection{Finding Legal Templates}
We can figure out which strings of featlets
constitute a template
(most sequences don't make sense, like {\small [\textsc{ArgSpan}, \textsc{Shape}]})
by brute force search with a few heuristics.
We have a rules which filter out strings of featlets like:
\begin{itemize}
\itemsep0em
\item nothing can come before a token extractor
\item if apply a featlet fails or doesn't change the context, stop there
	(this and all suffixes are invalid)
\end{itemize}

We do a breadth first search over all strings of featlets up to length 6 and
collect all strings which are templates: those that produce output on at
least 2 of 50 instances, producing 5241 templates.

At this point note that since featlets are functions from one context to
another, they are closed under function composition.

\subsection{Frequency-based Template Transforms}
For every template found, we
produce 5 additional templates by appending the following featlets:
\textsc{Top10}, \textsc{Top100}, \textsc{Top1000}, \textsc{Cnt8} \textsc{Cnt16}.
The \textsc{TopN} transforms a template by sorting its
features by frequency, and only letting the template fire for values with
a count at least as high as the $N^{th}$ most common feature.
\textsc{CntC} only lets through features observed at
least C times in the training data.

These automatic transforms are useful for building products, since they
control number of features created.
In our experiments, we found that a \textsc{TopN} template transform appeared in
a little less than 50\% of our final features and a \textsc{CntC} feature appeared
in a little less than 10\%.

\subsection{Template Composition}
To grow features larger than we can discover with brute force
enumeration of featlet strings, we consider products of templates.
It is common to represent this by string concatenation, but we
define template products to be the same as featlet concatenation.
This has one importance difference:
a template may return no value, in which case the rest of the product
returns no value.
With string concatenation you can represent one template making another
more specific by including more information.
With featlet composition you can have a template
modulate when another can fire.
This is weaker than general featlet composition though,
and order doesn't matter.

\subsection{Near Duplicate Removal}
\label{sec:near-dups}
We will generate pairs of similar and possibly redundant templates.
For example:
e.g. $X_1=$ {\small [\textsc{ArgHead}, \textsc{LeftSibD}, \textsc{Word}]}
and  $X_2=$ {\small [\textsc{ArgHead}, \textsc{LeftMostSibD}, \textsc{Word}]}.
Principled approaches like conditional mutual information $I(Y ; X_2 \mid X_1)$
could be used to filter, but this would require a lot of computation.
It is faster to use the type level (featlet/template names)
rather than the token level (values extracted on instance data).
We can construct similarity functions for each of the levels of structure we've produced.
\begin{enumerate}
\item similarity between two featlets is the normalized Levenshtein edit distance\footnote{Operations
have unit cost and we divide by the length of the longer string.} between their names,
so \textsc{LeftSibD} is similar to \textsc{LeftMostSibD} but not \textsc{ParentC}.
\item similarity between two templates (strings of featlets) is again the the normalized Levenshtein
edit distance, but over an alphabet of featlets, where the substitution cost
is inversely related to the previous similarity function.\footnote{We convert
edit distance to similarity by $sim(a,b) = \frac{k}{k + dist(a,b)}$ with $k=2$.}
\item similarity between two features is the max-weight bipartite matching of templates,
where the weight is inversely related to the previous similarity function.
We don't use edit distance here since order doesn't matter. 
\end{enumerate}
We use this last similarity function to prune ranked lists
of features produced in \S \ref{sec:feature-selection}.\footnote{We consider two
features redundant if the normalized max-weight matching is greater than 0.75.
To normalize we divide by the shorter length (in templates) of the two features.}


\section{Feature Selection}
\label{sec:feature-selection}

Feature generation can lead to too many features to fit in memory.
Some of the features we generate may provide no signal towards a label
we are trying to predict.
To filter down to a manageable set of informative features,
we score each template using mutual information (sometimes referred to
as information gain), between a label (Bernoulli) and a template (Multinomial).
Mutual information has a natural connection to Bayesian
independence testing \cite{minka2003bayesian}.
Since computing mutual information is just counting, this task
is embarrassingly parallel and can be easily implemented in
frameworks like MapReduce. 

We select a budget of how many features to
search over $B$, and divide that budget up amongst
template products up to order $n$ such that order $i$
features get a proportion of the budget of $\gamma^i$.
In these experiments we set $B=3000000$ and $\gamma=1.5$,
which meant that the split between features was
[21\%, 32\%, 47\%].\footnote{These are really maximum
proportions, filled up from lowest order to highest order,
with extra slots rolled over to the remaining slices proportional
to the remaining weights.}
% ... 
For each split, features were ranked by the max of a heuristic score
for each of its templates. Each templates heuristic score was its
mutual information plus a Gaussian with mean 0 and standard deviation 2.
Randomness was introduced for diversity and so that templates which
are more useful as filters (and have low mutual information by themselves)
have some chance of being picked.

Entropy and mutual information estimation breaks down
when the cardinality of the variables is large compared
to the number of observations \cite{paninski2003estimation}.
We observed that entropy estimates based on
the maximum likelihood estimates of $p(y,x)$ and $p(x)$
from counts of $(y,x)$ yielded very biased estimates of
mutual information (high for sparse features).
We correct for this problem by using the BUB entropy estimation
method described by \newcite{paninski2003estimation}.

\begin{figure}
\includegraphics[width=\linewidth]{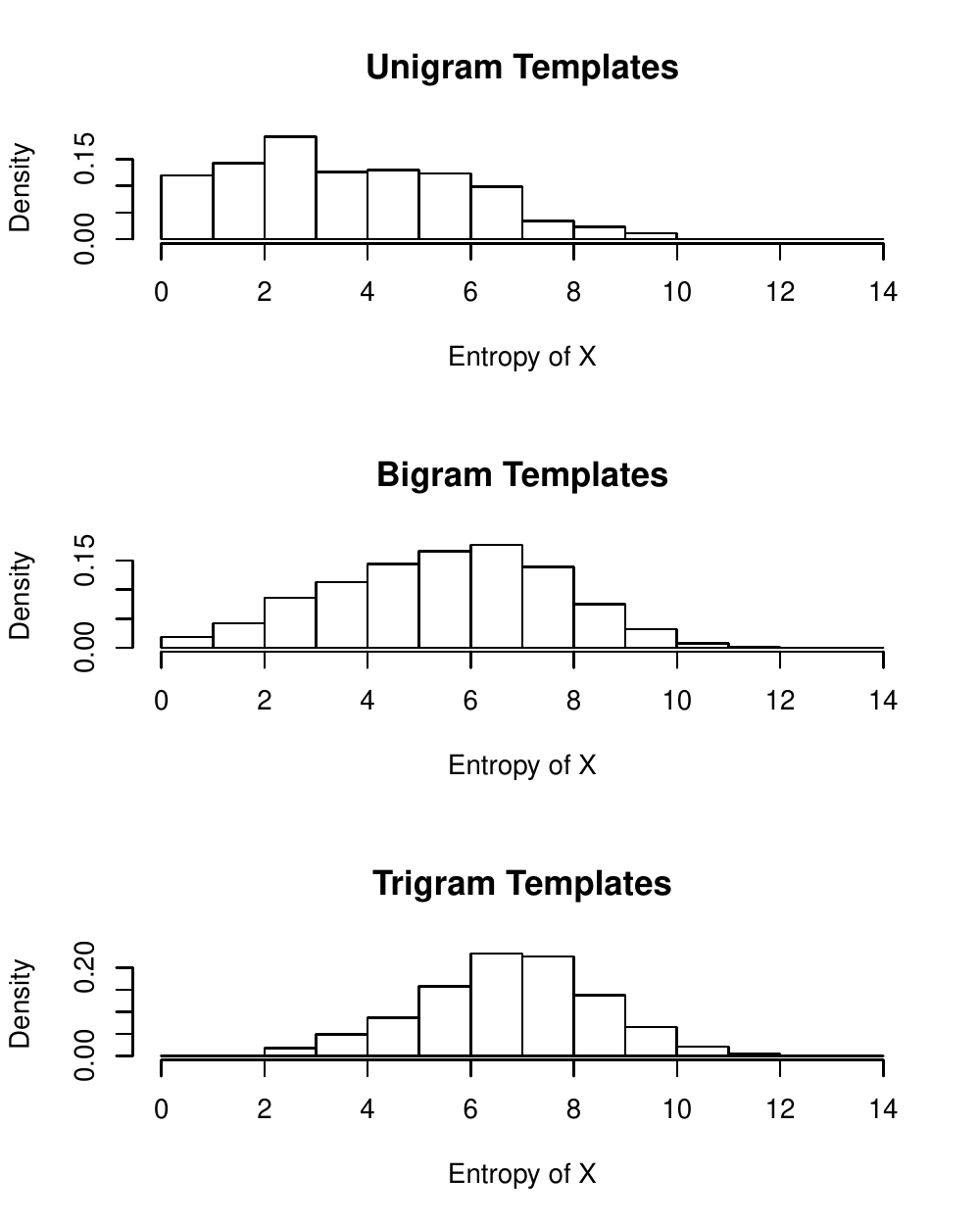}
\label{fig:hx}
\vspace{-1cm}
\caption{Entropy of template n-grams considered, in nats}
\end{figure}

We produce a final ranked list of features by sorting by
$\frac{I(Y;X)}{1 + \beta H(X)}$ and then applying the greedy
pruning described in \S \ref{sec:near-dups}.
This expression's limit as $\beta \rightarrow 0$ is mutual information
and normalized mutual information
as $\beta \rightarrow \infty$.
In our experiments, most features had between one and nine nats
of entropy, as shown in figure \ref{fig:hx},
and we created feature sets out of $\beta \in \{ 0.01, 0.1, 1, 10 \}$

\section{Experiments}
In our experiments we use semantic role labeling (SRL) as a
case study to test whether our automatic methods
can derive feature sets which work as well as hand engineered ones.
SRL is a difficult prediction task with more than one structural
formulation (type of label).  Sometimes arguments are represented by their head
token in a dependency tree \cite{surdeanu2008conll,hajivc2009conll}
and sometimes they are specified by
a span or constituent \cite{carreras2005,baker1998berkeley}.
For span-based SRL, the correspondence between the argument spans and
syntactic constituents can be very tight \cite{kingsbury2002treebank} or not \cite{ruppenhofer2006framenet}.
Sometimes the role labels depend on the predicate sense \cite{ruppenhofer2006framenet}
and sometimes they don't \cite{kingsbury2002treebank}.
These differences indicate that there may not be one ``SRL feature set" which
works best, or even well, for all variants of the task.



We used FrameNet 1.5 \cite{ruppenhofer2006framenet} and the
CoNLL 2012 data derived from the OntoNotes 5.0 corpus \cite{pradhan2013towards}.
We used the argument candidate selection method described in
\newcite{xue2004calibrating} as well as the extensions in \newcite{acl2014frames}.
Annotations are provided from the Stanford CoreNLP toolset \cite{manning-EtAl:2014:P14-5}.
Feature selection is run first on each data set to produce a
few feature sets based on $\beta$ and size, then we evaluate their
performance
using an averaged perceptron
\cite{freund1999large,collins2002discriminative}.  We ran the algorithm for up
to 10 passes over the data, shuffling before each pass, and selected the
averaged weights which yielded the best F1 on the dev set.\footnote{For
FrameNet data we took a random 10\% slice of the training data as the dev set.}

\paragraph{SRL Stages}
Most work on SRL breaks the problem into (at least) two stages: argument
identification and role classification.  The argument identification stage
classifies whether a span is a semantic argument to a particular target, and
then the classification stage chooses the role for each span which was selected
by the identification stage.
We adopt this standard architecture for efficiency: if there are $O(s)$ spans and
$O(k)$ roles, it turns an $O(s k)$ decoding problem into an $O(s + k)$
decoding problem.

Given this stage-based architecture, we split our budget,
half going to each stage.
For both we define $y$ to be a Bernoulli variable which is one on
sub-structures which appear on the gold parse.
For arg id the instances are spans for a particular target,
and for role classification the instances are roles for the span chosen
in the previous stage. During training we use gold arg id to
train the role classifier.
For arg id we only score features
which contain a \textsc{ArgHead}, or \textsc{ArgSpan} featlet,
and for role classification we additionally require
a \textsc{RoleArg} or \textsc{FrameRoleArg} featlet appear.

\section{Results}
Overall, our method seems to work about as well as experts manually
designing features for SRL.
Results in table \ref{tab:end2end} shows our approach matching the
performance of \newcite{Das:2012} and \newcite{pradhan2013towards}.
Other systems achieve better performance, but these models all
use global information, an orthogonal issue to the
local feature set.

\begin{table}
\resizebox{\columnwidth}{!}{
\begin{tabular}{l|r|r|r|r}
                                          & Global & P     & R     & F1          \\
\hline
This work                                 & \xmark & {\bf 73.9}  & 55.8  & 63.6        \\
\newcite{Das:2012} local                  & \xmark & 67.7  & {\bf 59.8}  & 63.5        \\
\newcite{Das:2012} constrained            & \cmark & 70.4  & 59.5  & {\bf 64.6}        \\
\hline
This work                                 & \xmark & {\bf 87.5}  & 69.1  & 77.2        \\
\newcite{pradhan2013towards}              & \xmark & 81.3  & 70.5  & 75.5        \\
\newcite{pradhan2013towards} (revised)    & \xmark & 78.5  & 76.7  & 77.5        \\
\newcite{tackstrom-etal-2015}             & \cmark & 80.6  & 78.2  & 79.4        \\
\newcite{fitzgerald:2015:EMNLP}           & \cmark & 80.9  & {\bf 78.4}  & 79.6  \\
\newcite{ZhouXu15}                        & \cmark & -     & -     & {\bf 81.3}  \\
\end{tabular}
}
\begin{small}
\caption{Performance of our automatic feature selection vs prior work.
In general our local model with automatic feature selection is
a few points behind joint inference models but matches or exceeds other local
inference models. Results for FrameNet are on top, Propbank below.}
\end{small}
\label{tab:end2end}
\end{table}

In looking at the feature sets generated,
one major difference is the complexity
parameter $\beta$. For FrameNet, the best value of $\beta$ was 10,
meaning that the features with the highest {\em normalized} mutual information
were chosen, whereas with Propbank and $\beta=0.01$,
it was better to ignore the entropy of the feature.
This makes sense in retrospect when you consider the size of the training sets,
Propbank is about 20 times larger, but its not clear how much data is needed
to justify this shift when tuning by hand.

This difference in selection criteria does lead to very different feature
sets chosen,\footnote{There are actually only two templates in common
between the best FrameNet and Propbank feature sets. Both contain the
\textsc{CommonParentD} featlet.}
but it is another question of whether this matters towards system performance.
It could be that there are many different types of feature sets which lead to
good performance on either task/dataset, and only one is needed
(possibly created manually).
In table \ref{table:cross-train} we show the effects generating a feature
set for one dataset and applying it to the other.
The performance on the diagonal is considerably higher, indicating empirically
that there likely isn't one ``SRL feature set".
If you weight both equally, the average
increase in error due to domain shift is 7.9\%.
This is even the case for feature selection with FrameNet, where you might
expect that selecting features on Propbank, a much larger resource, could
yield gains because of much lower variance without much bias.
\begin{table}
\begin{center}
\begin{tabular}{l | r r}
   & FN & PB \\
\hline
FN & 63.6 & 74.8 \\
PB & 61.7 & 77.2 \\
\end{tabular}
\label{table:cross-train}
\begin{small}
\caption{Columns are the dataset used for feature selection
and rows are the dataset used for training and testing.}
\end{small}
\end{center}
\end{table}

\paragraph{Sensitivity by Stage}
Given that we can automatically generate feature sets,
we can easily determine how adding or removing features
from each stage will affect performance.
This is useful for choosing a feature set which balances the cost
of prediction time with performance,
which is labor intensive and error prone when done manually.
Table \ref{table:pos-neg} shows that
the model is more sensitive to the removal of argument id features
than role classification ones.
This is not a new result \cite{Marquez:2008},
but this work offers a way to respond
by applying computational rather than human resources to the problem.
\begin{table}
\begin{center}
\begin{tabular}{r | r r r r}
FN   & 0 		& 320 	& 640 	& 1280 \\
\hline
0    & NA   & 50.1 	& 56.4 	& 61.5 \\
320  & 54.7 & 55.7 	& 58.8 	& 61.9 \\
640  & 57.8 & 59.6 	& 61.4 	& 62.4 \\
1280 & 58.1 & 59.5 	& 61.0 	& 63.6 \\
		 &			& 			& 			&      \\
PB   & 0 		& 320 	& 640 	& 1280 \\
\hline
0    & NA   & 59.9	& 65.8 	& 73.6 \\
320  & 61.3 & 62.8 	& 70.2 	& 74.5 \\
640  & 67.6 & 68.1 	& 74.1 	& 75.3 \\
1280 & 70.6 & 72.8 	& 75.1 	& 77.2 \\
\end{tabular}
\end{center}
\label{table:pos-neg}
\caption{Columns how many features were used for {\em argument identification}  
and rows how many features were used for {\em role classification}. 
FrameNet (FN) is on top $\beta=10$, Propbank (PB) is below $\beta=0.01$.}
\end{table}

\section{Discussion}
\paragraph{Limitations}
On limitation of the methods described here is finding symmetries.
The product operator for templates is commutative,
but this is not the case for featlets.
Some templates are equivalent and there is no easy way to
check short of checking their denotations,
which is expensive.

Another issue is that a lot of features are required.
The best models we trained for FrameNet use over 2500 features,
which is significantly more than
\newcite{Das:2012}, which used 34.
Upon manual inspection of the feature sets we learned,
we find most if not all of the features that \newcite{Das:2012}
created,\footnote{It is a not trivial to match our templates to
theirs. For example, the template
{\small
[\textsc{TargetHead}, \textsc{LeftL}, \textsc{Top10}]
* [\textsc{TargetHead}, \textsc{ChildSequenceD}, \textsc{SeqMapDepRel}, \textsc{Bag}]
}
is likely close to the passive voice template used in \newcite{Das:2012},
since ``was" and ``be" are in the top 10 words to the left of a verb.}
but precision is low.


\section{Related Work}
Recently there has been a swell in interest in
neural methods in NLP which use continuous representations
rather than discrete feature weights.
This work shares some motivation with neural methods,
e.g. the desire to avoid domain expert-derived features,
but we diverge primarily for computational reasons.
This work is about model generation and scoring,
and it is not clear how to score neural models
in ways that don't involve re-training a model.
Feature based models
are amenable decomposition and information theoretic analysis
in ways that neural models aren't.

Within feature based methods, backwards selection methods
are common,
including a large body of work on sparsity-inducing regularization,
the canonical being the lasso \cite{tibshirani1996}.
These methods are applicable when the entire feature set can be enumerated
and scored on one machine.
This is not feasible for this work,
since we generate
features which lie in a combinatorial space too large
to fit in memory.
For example, our method found the feature
{\small [
\textsc{TargetHead},
\textsc{Right},
\textsc{WnSynset},	
\textsc{ArgSpan},
\textsc{Span1Start},
\textsc{LeftL},
\textsc{Word},
\textsc{Top10},			
\textsc{Span1ToSpan2L},
\textsc{SeqMapPos},
\textsc{Bag}
]}, which is comprised of 11 featlets\footnote{Technically it is 13 featlets
since an \textsc{Output} featlet is not written after the \textsc{WnSynset}
and \textsc{Top10} featlets, \S \ref{sec:featlets-formal}.}.

The alternative are forwards selection methods which work
by building bigger features from smaller ones.
\newcite{Bjorkelund:2009} used forwards selection for dependency-based
SRL \cite{hajivc2009conll} for 7 languages based on products of templates.
Their experiments showed a great diversity of the features learned for different languages
and they placed second in the shared task.
\newcite{McCallum:2002} used forwards selection for named entity recognition,
scoring new feature products using approximate model re-fitting (pseudo-likelihood),
which also produced good results.
In both of these works, scoring new features depended on the output of a smaller
feature set.
Sequential methods like this are not amenable to paralellization and take
quadratic time with respect to the number of feature to be searched over.

Our method is more similar to the work of \newcite{gormley-etal:2014:SRL}
where every template is scored in parallel
irrespective of a trained model.





\paragraph{Future work}


This work dove-tails with the approach described by
\newcite{Lee:2007}, which derives a prior or regularization
constant for individual features by looking at properties of
the feature (meta features).
This work generates features with a lot of structure,
which the learner could reflect upon to improve regularization
and generalization.

The structure in these features can also inform parameterization.
Tensor decomposition methods of
fixed-order tensors have been used to great effect \cite{lei:2014,lei:2015}.
Low-rank or embedding methods (e.g. RNNs) for parameterizing
featlet strings, as opposed to storing a weight in a dictionary,
could also improve regularization.


Step-wise methods which select some features, fit a model, and
then select more features with respect to mutual information with
residuals, are another simple and promising direction.

\section{Conclusion}
In this work we propose a general framework for generating feature sets
with the goal of removing expert engineering from the machine learning loop.
Our approach is based on composing units called featlets to create templates.
Featlets are small functions which are task agnostic and easy to
define and implement by non-experts.
Featlets on one hand preserve a wide variety of nuanced feature semantics,
and on the other can be enumerated automatically to derive a huge amount of novel
templates and features.
We validate our approach on semantic role labeling and achieve performance
on par with models that had considerable expert intervention.

\bibliography{acl2016}{}

\begin{thebibliography}{}

\bibitem[\protect\citename{Baker \bgroup et al.\egroup
  }1998]{baker1998berkeley}
Collin~F Baker, Charles~J Fillmore, and John~B Lowe.
\newblock 1998.
\newblock The berkeley framenet project.
\newblock In {\em Proceedings of the 36th Annual Meeting of the Association for
  Computational Linguistics and 17th International Conference on Computational
  Linguistics-Volume 1}. ACL.

\bibitem[\protect\citename{Bj\"{o}rkelund \bgroup et al.\egroup
  }2009]{Bjorkelund:2009}
Anders Bj\"{o}rkelund, Love Hafdell, and Pierre Nugues.
\newblock 2009.
\newblock Multilingual semantic role labeling.
\newblock In {\em Proceedings of the Thirteenth Conference on Computational
  Natural Language Learning: Shared Task}, CoNLL '09, pages 43--48,
  Stroudsburg, PA, USA. Association for Computational Linguistics.

\bibitem[\protect\citename{Carreras and M{\`a}rquez}2005]{carreras2005}
Xavier Carreras and Llu{\'\i}s M{\`a}rquez.
\newblock 2005.
\newblock Introduction to the conll-2005 shared task: Semantic role labeling.
\newblock In {\em Proceedings of the Ninth Conference on Computational Natural
  Language Learning}, pages 152--164. Association for Computational
  Linguistics.

\bibitem[\protect\citename{Collins}2002]{collins2002discriminative}
Michael Collins.
\newblock 2002.
\newblock Discriminative training methods for hidden markov models: Theory and
  experiments with perceptron algorithms.
\newblock In {\em Proceedings of the ACL-02 conference on Empirical methods in
  natural language processing-Volume 10}, pages 1--8. Association for
  Computational Linguistics.

\bibitem[\protect\citename{Das \bgroup et al.\egroup }2012]{Das:2012}
Dipanjan Das, Andr{\'e} F.~T. Martins, and Noah~A. Smith.
\newblock 2012.
\newblock An exact dual decomposition algorithm for shallow semantic parsing
  with constraints.
\newblock In {\em SemEval}, SemEval '12. Association for Computational
  Linguistics.

\bibitem[\protect\citename{Das \bgroup et al.\egroup }2014]{DasFramesCL:2014}
Dipanjan Das, Desai Chen, André F.~T. Martins, Nathan Schneider, and Noah~A.
  Smith.
\newblock 2014.
\newblock Frame-semantic parsing.
\newblock {\em Computational Linguistics}, 40:1:9--56.

\bibitem[\protect\citename{FitzGerald \bgroup et al.\egroup
  }2015]{fitzgerald:2015:EMNLP}
Nicholas FitzGerald, Oscar T{\"{a}}ckstr{\"{o}}m, Kuzman Ganchev, and Dipanjan
  Das.
\newblock 2015.
\newblock Semantic role labelling with neural network factors.
\newblock In {\em Proceedings of the 2015 Conference on Empirical Methods in
  Natural Language Processing}, Lisboa, Portugal, September. Association for
  Computational Linguistics.

\bibitem[\protect\citename{Freund and Schapire}1999]{freund1999large}
Yoav Freund and Robert~E Schapire.
\newblock 1999.
\newblock Large margin classification using the perceptron algorithm.
\newblock {\em Machine learning}, 37(3):277--296.

\bibitem[\protect\citename{Gildea and Jurafsky}2002]{gildea2002automatic}
Daniel Gildea and Daniel Jurafsky.
\newblock 2002.
\newblock Automatic labeling of semantic roles.
\newblock {\em Computational linguistics}, 28(3).

\bibitem[\protect\citename{Gormley \bgroup et al.\egroup
  }2014]{gormley-etal:2014:SRL}
Matthew~R. Gormley, Margaret Mitchell, Benjamin {Van Durme}, and Mark Dredze.
\newblock 2014.
\newblock Low-resource semantic role labeling.
\newblock In {\em Proceedings of {ACL}}, June.

\bibitem[\protect\citename{Haji{\v{c}} \bgroup et al.\egroup
  }2009]{hajivc2009conll}
Jan Haji{\v{c}}, Massimiliano Ciaramita, Richard Johansson, Daisuke Kawahara,
  Maria~Ant{\`o}nia Mart{\'\i}, Llu{\'\i}s M{\`a}rquez, Adam Meyers, Joakim
  Nivre, Sebastian Pad{\'o}, Jan {\v{S}}t{\v{e}}p{\'a}nek, et~al.
\newblock 2009.
\newblock The conll-2009 shared task: Syntactic and semantic dependencies in
  multiple languages.
\newblock In {\em Proceedings of the Thirteenth Conference on Computational
  Natural Language Learning: Shared Task}, pages 1--18. Association for
  Computational Linguistics.

\bibitem[\protect\citename{Hermann \bgroup et al.\egroup }2014]{acl2014frames}
Karl~Moritz Hermann, Dipanjan Das, Jason Weston, and Kuzman Ganchev.
\newblock 2014.
\newblock Semantic frame identification with distributed word representations.
\newblock In {\em Proceedings of ACL}. Association for Computational
  Linguistics.

\bibitem[\protect\citename{Johansson and Nugues}2008]{Johansson:2008}
Richard Johansson and Pierre Nugues.
\newblock 2008.
\newblock Dependency-based semantic role labeling of propbank.
\newblock In {\em Proceedings of the Conference on Empirical Methods in Natural
  Language Processing}, EMNLP '08, pages 69--78, Stroudsburg, PA, USA.
  Association for Computational Linguistics.

\bibitem[\protect\citename{Kingsbury and Palmer}2002]{kingsbury2002treebank}
Paul Kingsbury and Martha Palmer.
\newblock 2002.
\newblock From treebank to propbank.
\newblock In {\em LREC}. Citeseer.

\bibitem[\protect\citename{Lee \bgroup et al.\egroup }2007]{Lee:2007}
Su-In Lee, Vassil Chatalbashev, David Vickrey, and Daphne Koller.
\newblock 2007.
\newblock Learning a meta-level prior for feature relevance from multiple
  related tasks.
\newblock In {\em Proceedings of the 24th International Conference on Machine
  Learning}, ICML '07, pages 489--496, New York, NY, USA. ACM.

\bibitem[\protect\citename{Lei \bgroup et al.\egroup }2014]{lei:2014}
Tao Lei, Yu~Xin, Yuan Zhang, Regina Barzilay, and Tommi Jaakkola.
\newblock 2014.
\newblock Low-rank tensors for scoring dependency structures.
\newblock In {\em Proceedings of the 52nd Annual Meeting of the Association for
  Computational Linguistics (Volume 1: Long Papers)}, pages 1381--1391,
  Baltimore, Maryland, June. Association for Computational Linguistics.

\bibitem[\protect\citename{Lei \bgroup et al.\egroup }2015]{lei:2015}
Tao Lei, Yuan Zhang, Llu\'{i}s M\`{a}rquez, Alessandro Moschitti, and Regina
  Barzilay.
\newblock 2015.
\newblock High-order low-rank tensors for semantic role labeling.
\newblock In {\em Proceedings of the 2015 Conference of the North American
  Chapter of the Association for Computational Linguistics: Human Language
  Technologies}, pages 1150--1160, Denver, Colorado, May--June. Association for
  Computational Linguistics.

\bibitem[\protect\citename{Liang}2005]{liang2005semi}
Percy Liang.
\newblock 2005.
\newblock Semi-supervised learning for natural language.
\newblock Master's thesis, Massachusetts Institute of Technology.

\bibitem[\protect\citename{Manning \bgroup et al.\egroup
  }2014]{manning-EtAl:2014:P14-5}
Christopher~D. Manning, Mihai Surdeanu, John Bauer, Jenny Finkel, Steven~J.
  Bethard, and David McClosky.
\newblock 2014.
\newblock The {Stanford} {CoreNLP} natural language processing toolkit.
\newblock In {\em Association for Computational Linguistics (ACL) System
  Demonstrations}, pages 55--60.

\bibitem[\protect\citename{M\`{a}rquez \bgroup et al.\egroup
  }2008]{Marquez:2008}
Llu\'{\i}s M\`{a}rquez, Xavier Carreras, Kenneth~C. Litkowski, and Suzanne
  Stevenson.
\newblock 2008.
\newblock Semantic role labeling: An introduction to the special issue.
\newblock {\em Comput. Linguist.}, 34(2):145--159, June.

\bibitem[\protect\citename{McCallum}2003]{McCallum:2002}
Andrew McCallum.
\newblock 2003.
\newblock Efficiently inducing features of conditional random fields.
\newblock In {\em Proceedings of the Nineteenth Conference on Uncertainty in
  Artificial Intelligence}, UAI'03, San Francisco, CA, USA. Morgan Kaufmann
  Publishers Inc.

\bibitem[\protect\citename{Minka}2003]{minka2003bayesian}
Tom Minka.
\newblock 2003.
\newblock Bayesian inference, entropy, and the multinomial distribution.
\newblock {\em Online tutorial}.

\bibitem[\protect\citename{Paninski}2003]{paninski2003estimation}
Liam Paninski.
\newblock 2003.
\newblock Estimation of entropy and mutual information.
\newblock {\em Neural computation}, 15(6):1191--1253.

\bibitem[\protect\citename{Pradhan \bgroup et al.\egroup
  }2005]{pradhan2005support}
Sameer Pradhan, Kadri Hacioglu, Valerie Krugler, Wayne Ward, James~H Martin,
  and Daniel Jurafsky.
\newblock 2005.
\newblock Support vector learning for semantic argument classification.
\newblock {\em Machine Learning}, 60(1-3):11--39.

\bibitem[\protect\citename{Pradhan \bgroup et al.\egroup
  }2013]{pradhan2013towards}
Sameer Pradhan, Alessandro Moschitti, Nianwen Xue, Hwee~Tou Ng, Anders
  Bj{\"o}rkelund, Olga Uryupina, Yuchen Zhang, and Zhi Zhong.
\newblock 2013.
\newblock Towards robust linguistic analysis using ontonotes.
\newblock In {\em Proceedings of the Seventeenth Conference on Computational
  Natural Language Learning. Sofia, Bulgaria: Association for Computational
  Linguistics}, pages 143--52.

\bibitem[\protect\citename{Punyakanok \bgroup et al.\egroup
  }2008]{punyakanok2008importance}
Vasin Punyakanok, Dan Roth, and Wen-tau Yih.
\newblock 2008.
\newblock The importance of syntactic parsing and inference in semantic role
  labeling.
\newblock {\em Computational Linguistics}, 34(2):257--287.

\bibitem[\protect\citename{Ruppenhofer \bgroup et al.\egroup
  }2006]{ruppenhofer2006framenet}
Josef Ruppenhofer, Michael Ellsworth, Miriam~RL Petruck, Christopher~R Johnson,
  and Jan Scheffczyk.
\newblock 2006.
\newblock Framenet ii: Extended theory and practice.

\bibitem[\protect\citename{Surdeanu \bgroup et al.\egroup
  }2008]{surdeanu2008conll}
Mihai Surdeanu, Richard Johansson, Adam Meyers, Llu{\'\i}s M{\`a}rquez, and
  Joakim Nivre.
\newblock 2008.
\newblock The conll-2008 shared task on joint parsing of syntactic and semantic
  dependencies.
\newblock In {\em Proceedings of the Twelfth Conference on Computational
  Natural Language Learning}, pages 159--177. Association for Computational
  Linguistics.

\bibitem[\protect\citename{T{\"a}ckstr{\"o}m \bgroup et al.\egroup
  }2015]{tackstrom-etal-2015}
Oscar T{\"a}ckstr{\"o}m, Kuzman Ganchev, and Dipanjan Das.
\newblock 2015.
\newblock Efficient inference and structured learning for semantic role
  labeling.
\newblock {\em Transactions of the Association for Computational Linguistics},
  3.

\bibitem[\protect\citename{Tibshirani}1996]{tibshirani1996}
Robert Tibshirani.
\newblock 1996.
\newblock Regression shrinkage and selection via the lasso.
\newblock {\em Journal of the Royal Statistical Society. Series B
  (Methodological)}, pages 267--288.

\bibitem[\protect\citename{Toutanova \bgroup et al.\egroup
  }2005]{Toutanova:2005}
Kristina Toutanova, Aria Haghighi, and Christopher~D. Manning.
\newblock 2005.
\newblock Joint learning improves semantic role labeling.
\newblock In {\em Proceedings of the 43rd Annual Meeting on Association for
  Computational Linguistics}, ACL '05. Association for Computational
  Linguistics.

\bibitem[\protect\citename{Xue and Palmer}2004]{xue2004calibrating}
Nianwen Xue and Martha Palmer.
\newblock 2004.
\newblock Calibrating features for semantic role labeling.
\newblock In {\em EMNLP}, pages 88--94.

\bibitem[\protect\citename{Zhou and Xu}2015]{ZhouXu15}
Jie Zhou and Wei Xu.
\newblock 2015.
\newblock End-to-end learning of semantic role labeling using recurrent neural
  networks.
\newblock In {\em Proceedings of the 53rd Annual Meeting of the Association for
  Computational Linguistics and the 7th International Joint Conference on
  Natural Language Processing of the Asian Federation of Natural Language
  Processing, {ACL} 2015, July 26-31, 2015, Beijing, China, Volume 1: Long
  Papers}, pages 1127--1137.

\end{thebibliography}
\bibliographystyle{acl2016}

\end{document}